\documentclass[11pt,a4paper]{article}
\usepackage[hyperref]{acl2017}
\usepackage{times}
\usepackage{graphicx}
\usepackage{subfigure}
\usepackage{tabularx}
\usepackage{booktabs}

\usepackage{latexsym}
\usepackage{microtype}
\usepackage{multirow}

\usepackage{amssymb}
\usepackage{tipa}
\usepackage{amsmath}
\usepackage{covington}

\aclfinalcopy

\def\aclpaperid{***}

\title{Decoding Sentiment from Distributed Representations of Sentences}

\author{Edoardo Maria Ponti \\ {\tt ep490@cam.ac.uk} \And Ivan Vuli\'c \\ {\tt iv250@cam.ac.uk} \\\\ Language Technology Lab, University of Cambridge \And Anna Korhonen\\
   {\tt alk23@cam.ac.uk}}
   
\date{}

\begin{document}

\maketitle

\begin{abstract}
Distributed representations of sentences have been developed recently to represent their meaning as real-valued vectors. However, it is not clear how much information such representations retain about the polarity of sentences. To study this question, we decode sentiment from unsupervised sentence representations learned with different architectures (sensitive to the order of words, the order of sentences, or none) in 9 typologically diverse languages. Sentiment results from the (recursive) composition of lexical items and grammatical strategies such as negation and concession. The results are manifold: we show that there is no `one-size-fits-all' representation architecture outperforming the others across the board. Rather, the top-ranking architectures depend on the language and data at hand. Moreover, we find that in several cases the additive composition model based on skip-gram word vectors may surpass supervised state-of-art architectures such as bidirectional LSTMs. Finally, we provide a possible explanation of the observed variation based on the type of negative constructions in each language. 

\end{abstract}

\section{Introduction}


Distributed representations of sentences are usually acquired in an unsupervised fashion from raw texts. Those inferred from different algorithms are prone to grasp parts of their meaning and disregard others. Representations have been evaluated thoroughly, both intrinsically (interpretation through distance measures) and extrinsically (performance on downstream tasks). Moreover, several methods have been considered, based on both the composition of word embeddings \cite{milajevs10evaluating,marelli2014semeval,sultan2015dls} and direct generation \cite{hill2016learning}. The evaluation was focused solely on English, and it rarely concerned other languages \cite{Adi:2017iclr,Conneau:2017arxiv}. As a consequence, many `core' methods to learn distributed sentence representations are largely under-explored in a variety of typologically diverse languages, and still lack a demonstration of their usefulness in actual downstream tasks.

In this work, we study how well distributed sentence representations capture the \textit{polarity of a sentence}. To this end, we choose the Sentiment Analysis task as an extrinsic evaluation protocol: it directly detects the polarity of a text, where polarity is defined as the attitude of the speaker with respect to the whole content of the string or one of the entities mentioned therein. This attitude is measured quantitatively on a scale spanning from negative to positive with arbitrary granularity. As such, polarity consists in a crucial part of the meaning of a sentence, which should not be lost.

The polarity of a sentence depends heavily on a complex interaction between lexical items endowed with an intrinsic polarity, and morphosyntactic constructions altering polarity, most notably negation and concession.
The interaction is deemed to be recursive, hence some approaches take into account word order and phrase boundaries in order to apply the correct composition
\cite{socher2013recursive}.
However, some languages lack continuous constituents: contiguous spans of words do not correspond to syntactic subtrees, making composition unreliable \cite{ponti2016divergence}.
Moreover, the expression of negation varies across languages, as demonstrated by works in Linguistic Typology \cite[\textit{inter alia}]{dahl1979typology}. In particular, negation can appear as a bounded morpheme or a free morpheme; it can precede or follow the verb; it can `agree' or not in polarity with indefinite pronouns; it can alter the expression of verbal categories (e.g.\ tense, aspect, or modality).

We explore a series of methods endowed with different features: some hinge upon word order, others on sentence order, others on neither. We evaluate these unsupervised representations using a Multi-Layer Perceptron which uses the generated sentence representations as input and predicts sentiment classes (positive vs. negative) as output. Training and evaluation are based on a collection of annotated databases. Owing to the variety of methods and languages, we expect to observe a variation in the performance correlated with the properties of both.



Moreover, we establish a ceiling to the possible performances of our method based on decoding unsupervised distributed representations. In fact, we offer a comparison between this and supervised deep learning architectures that achieve state-of-art scores in the Sentiment Analysis task. In particular, we also evaluate a bi-directional LSTM \cite{li2015visualizing} on the same task. These models have advantage over distributed representations as: i) they are specialised on a single task rather than built as general-purpose representations; ii) their recurrent nature allows to capture the sequential composition of polarity in a sentence. However, since training these models requires large amounts of annotated data, resource scarcity in other languages hampers their portability.



The aim of this work is to assess which algorithm for distributed sentence representations is the most appropriate for capturing polarity in a given language. Moreover, we study how language-specific properties have an impact on performance, finding an explanation in Language Typology. We also provide an in-depth analysis of the most relevant features by visualising the activation of hidden neurons. This will hopefully contribute to advancing the Sentiment Analysis task in the multilingual scenarios. In \S\ \ref{sec:msa}, we survey prior work on multilingual sentiment analysis. Afterwards, we present the tested algorithms for generating distributed representations of sentences in \S\ \ref{sec:algorithms}. In \S\ \ref{sec:setup}, we sketch the dataset and the experimental setup. Finally, \S\ \ref{sec:results} examines the results in light of the sensitivity of the algorithms and the typology of negation.

\section{Multilingual Sentiment Analysis}
\label{sec:msa}

The task of sentiment classification is mostly addressed through supervised approaches. However, these achieve unsatisfactory results in resource-lean languages because of the scarcity of resources to train dedicated models \cite{denecke2008using}. This afflicts state-of-art deep learning architectures even more compared to traditional machine learning algorithms \cite{chen2016adversarial}. As a consequence, previous work resorted to i) language transfer or ii) joint multilingual learning. The former adapts models from a source resource-rich language to a target resource-poor language; the latter infers a single model portable across languages. Approaches based on distributed representations induced in an unsupervised fashion do not face the difficulty resulting from resource scarcity: they are portable to other tasks and languages. In this section we survey deep learning techniques, adaptive models, and unsupervised distributed representations for sentiment classification in a multilingual scenario. The last approach is the focus of this work.

Deep learning algorithms for sentiment classification are designed to deal with compositionality. Hence, they often rely on recurrent networks tracing the sequential history of a sentence, or special compositional devices. Recurrent models include bi-directional LSTMs \cite{li2015visualizing}, possibly enriched with context \cite{mousacontextual}.
On the other hand, \newcite{socher2013recursive} put forth a Recursive Neural Tensor Network, which composes representations recursively through a single tensor-based composition function. Subsequent improvements of this line of research include the Structural Attention Neural Networks \cite{kokkinos2017structural}, which adds  structural information around each
node of a syntactic tree.

When supervised monolingual models are not feasible, language transfer can bridge between multiple languages, for instance through supervised latent Dirichlet allocation \cite{boyd2010holistic}. Direct transfer relies on word-aligned parallel texts where the source language text is either manually or automatically annotated. The sentiment information is then projected onto the target text \cite{almeida2015aligning}, also leveraging non-parallel data \cite{zhou2015subspace}. \newcite{chen2016adversarial} devised a multi-task network where an adversarial branch spurs the shared layers to learn language-independent features. Finally, \newcite{lu2011joint} learned from annotated examples in both the source and target language. Alternatively, sentences from other languages are translated into English and assigned a sentiment based on lexical resources \cite{denecke2008using} or supervised methods \cite{balahur2014comparative}.

Finally, cross-lingual sentiment classification can leverage on shared distributed representations. \newcite{zhou2016cross} captured shared high-level features across aligned sentences through autoencoders. In this latent space, distances were optimised to reflect differences in sentiment. On the other hand, \newcite{fernandez2015distributional} exploited bilingual word representations, where vector dimensions mirror the distributional overlap with respect to a pivot. \newcite{le2014distributed} concatenated sentence representations obtained through variants of Paragraph Vector and trained a Logistic Regression model on top of them.


Previous studies thus demonstrated that sentence representations retain information about polarity, and that they partly alleviate the drawbacks of deep architectures (single-purposed and data-demanding). Hence, the Sentiment Analysis tasks seems convenient to compare different sentence representation architectures. Nonetheless, a systematic evaluation has never taken place for this task, and a large-scale study over typologically diverse languages has not been attempted for any of the algorithms reviewed. We intend to fill these gaps, considering the methods to generate sentence representations outlined in the next section.

\section{Distributed Sentence Representations}
\label{sec:algorithms}


Word vectors can be combined through various compositional operations to obtain representations of phrases and sentences. \newcite{mitchell2010composition} explored two operations: addition and multiplication. Notwithstanding their simplicity, they are hardly outperformed by more sophisticated operations \cite{rimell2016relpron}. Some of these compositional representations based on matrix multiplication were also evaluated on sentiment classification \cite{yessenalina2011compositional}. Alternatively, sentence representations can be induced directly with no intermediate step at the word level. In this paper, we focus on sentence representations that are generated in an unsupervised fashion. Furthermore, they are `fixed', that is, they are not fine-tuned for any particular downstream task, since we are interested in their intrinsic content.\footnote{This excludes methods concerned with phrases, like the ECO embeddings \cite{poliak2017efficient}, or requiring structured knowledge, like CHARAGRAM \cite{wieting2016charagram}.}

\subsection{Algorithms}
We explore several methods to generate sentence representations. One exploits a compositional operation (addition) over word representations stemming from a Skip-Gram model (\S\ \ref{sssec:skipgram}). Others are direct methods, including FastSent (\S\ \ref{sssec:fastsent}), a Sequential Denoising AutoEncoder (SDAE, \S\ \ref{sssec:sdae}) and Paragraph Vector (\S\ \ref{sssec:doc2vec}). Note that FastSent relies on sentence order, SDAE on word order, and Paragraph Vector on neither. All these algorithms were trained on cleaned-up Wikipedia dumps.

The choice of the algorithms was based on following criteria: i) their performance reported in recent surveys (n.b., the surveys were limited to English and evaluated on other tasks), most notably \newcite{hill2016learning} and \newcite{milajevs10evaluating}; ii) the variety of their modelling assumptions and features encoded. The referenced surveys already hinted that the usefulness of a representation is largely dependent on the actual application. Shallower but more interpretable representations can be decoded with spatial distance metrics. Others, more deep and convoluted architectures, outperform the others in supervised tasks. We inquire whether the generalisation is tenable also in the task of Sentiment Analysis targeting sentence polarity.

\subsubsection{Additive Skip-Gram}
\label{sssec:skipgram}
As a bottom-up method, we train word embeddings using skip-gram with negative sampling \cite{mikolov2013distributed}. The algorithm finds the parameter $\theta$ such that, given a pair of a word $w$ and a context $c$, the model discriminates correctly whether it belongs to a set of sentences $S$ or a set of randomly generated incorrect sentences $S'$:

\begin{align}
\prod_{(w, c) \in S} p(S = 1 | w,c,\theta) \prod_{(w, c) \in S'} p(S' = 0 | w,c,\theta) \notag
\end{align}
The representation of a sentence was obtained via element-wise addition of the vectors of the words belonging to it \cite{mitchell2010composition}.


\subsubsection{FastSent}
\label{sssec:fastsent}

The FastSent model was proposed by \newcite{hill2016learning}. It hinges on a sentence-level distributional hypothesis \cite{polajnar2015exploration,kiros2015skip}. In other terms, it assumes that the meaning of a sentence can be inferred by the neighbour sentences in a text. It is a simple additive log-linear model conceived to mitigate the computational expensiveness of algorithms based on a similar assumption. Hence, it was preferred over SkipThought \cite{kiros2015skip} because of i) these efficiency issues and ii) its competitive performances reported by \newcite{hill2016learning}. In FastSent, sentences are represented as bags of words: a context of sentences is used to predict the adjacent sentence. Each word \textit{w} corresponds to a source vector $u_w$ and a target vector $v_w$. A sentence $S_i$ is represented as the sum of the source vectors of its words $\sum_{w \in S_i} u_w$. Hence, the cost \textit{C} of a representation is given by the softmax $\sigma(x)$ of a sentence representation and the target vectors of the words in its context \textit{c}.

\begin{equation}
C_{S_i} = \sum_{c \in S_{i-1} \cup S_{i+1}} \sigma(\sum_{w \in S_i} u_w, v_c)
\end{equation}
This model does not rely on word order, but rather on sentence order. It encodes new sentences by summing over the source vectors of their words.

\subsubsection{Sequential Denoising AutoEncoder}
\label{sssec:sdae}

Sequential Denoising AutoEncoders (SDAEs) combine features of Denoising AutoEncoders (DAE) and Sequence-to-Sequence models. In DAE, the input representation is corrupted by a noise function and the algorithms learns to recover the original \cite{vincent2008extracting}. Intuitively, this makes the model more robust to changes in input that are irrelevant for the task at hand. This architecture was later adapted to encode and decode variable-length inputs, and the corruption process was implemented in the form of dropout \cite{iyyer2015deep}. In the implementation by \newcite{hill2016learning},\footnote{\url{https://github.com/fh295/SentenceRepresentation}} the corruption function is defined as $f(S|p_o, p_x)$. $S$ is a list of words (a sentence) where each has a probability $p_o$ to be deleted, and the order of the words in every distinct bigram has a probability $p_x$ to be swapped. The architecture consists in a Recurrent Layer and predicts $p(S|f(S|p_o, p_x))$.

\subsubsection{Paragraph Vector}
\label{sssec:doc2vec}

Paragraph Vector is a collection of log-linear models proposed by \newcite{le2014distributed} for paragraph/sentence representation. It consists of two different models, namely the Distributed Memory model (DM) and the Distributed Bag Of Words model (DBOW). In DM, the ID of every distinct paragraph (or sentence) is mapped to a unique vector in a matrix \textit{D} and each word is mapped to a unique vector in matrix \textit{W}. Given a sentence \textit{i} and a window size \textit{k}, the vector $D_{i,\cdot}$ is used in conjunction with the concatenation of the vectors of the words in a sampled context $\left<w_{i_1},\dots,w_{i_k}\right>$ to predict the next word through logistic regression:

\begin{equation}
p(W_{i_{k+1}} | \left<D_i, W_{i_1},\dots, W_{i_k}\right>)
\end{equation}
Note that the sentence ID vector is shared by the contexts sampled from the same sentence. On the other hand, DBOW focuses on predicting the word embedding $W_{i_j}$ for a sampled word \textit{j} belonging to sentence \textit{i} given the sentence representation $D_i$. As a result, the main difference between the two Paragraph Vector models is that the first is sensitive to word order (represented by the word vector concatenation), whereas the second is insensitive with respect to it. These models store a representation for each sentence in the training set, hence they are memory demanding. We use the \textit{gensim} implementation of the two models available as Doc2Vec.\footnote{\url{https://radimrehurek.com/gensim/models/doc2vec.html}}

\subsection{Hyper-parameters}
The choice of the models' hyper-parameters was based on two (contrasting) criteria: i) conservativeness with those proposed in the original models and ii) comparability among the models in this work. In particular, we ensured that each model had the same sentence vector dimensionality: 300. The only exception is SDAE: we kept the recommended value of 2400. Paragraph Vector DBOW and SkipGram were trained for 10 epochs, with a window size of 10, a minimum frequency count of 5, and a sampling threshold of $10^{-5}$. FastSent was set as having a minimum count of 3, and no sampling. The probabilities in the corruption function of the SDAE were set as $p_o = 0.1$ (deletion) and $p_x = 0.1$ (swapping). The dimension of the RNN (GRU) hidden states (and hence sentence vector) was 2400, whereas single words were assigned 100 dimensions. The learning rate was set to 0.01 without decay, and the training lasted 7.2 hours on a \textsc{nvidia} Titan X \textsc{gpu}. The main properties of each algorithm are summarised in Table~\ref{tab:hyperparams}.

\begin{table}[ht]
\centering
{\footnotesize
\begin{tabular}{|l|c|c|}
\hline
\textbf{Algorithm} &	\textbf{WO} &	\textbf{SO}	 \\
\hline
Additive SkipGram & & \\
\hline
ParagraphVec DBOW & &\\
\hline
FastSent & & $\checkmark$ \\ 
\hline
Sequential Denoising AutoEncoder & $\checkmark$&\\
\hline
\end{tabular}
}%
\caption{Sensitivity to Word or Sentence Order.}
\label{tab:hyperparams}
\end{table}

\begin{figure*}[th]
     \begin{center}
        \subfigure[Arabic]{%
            \label{fig:neg_ar}
            \includegraphics[width=0.24\textwidth]{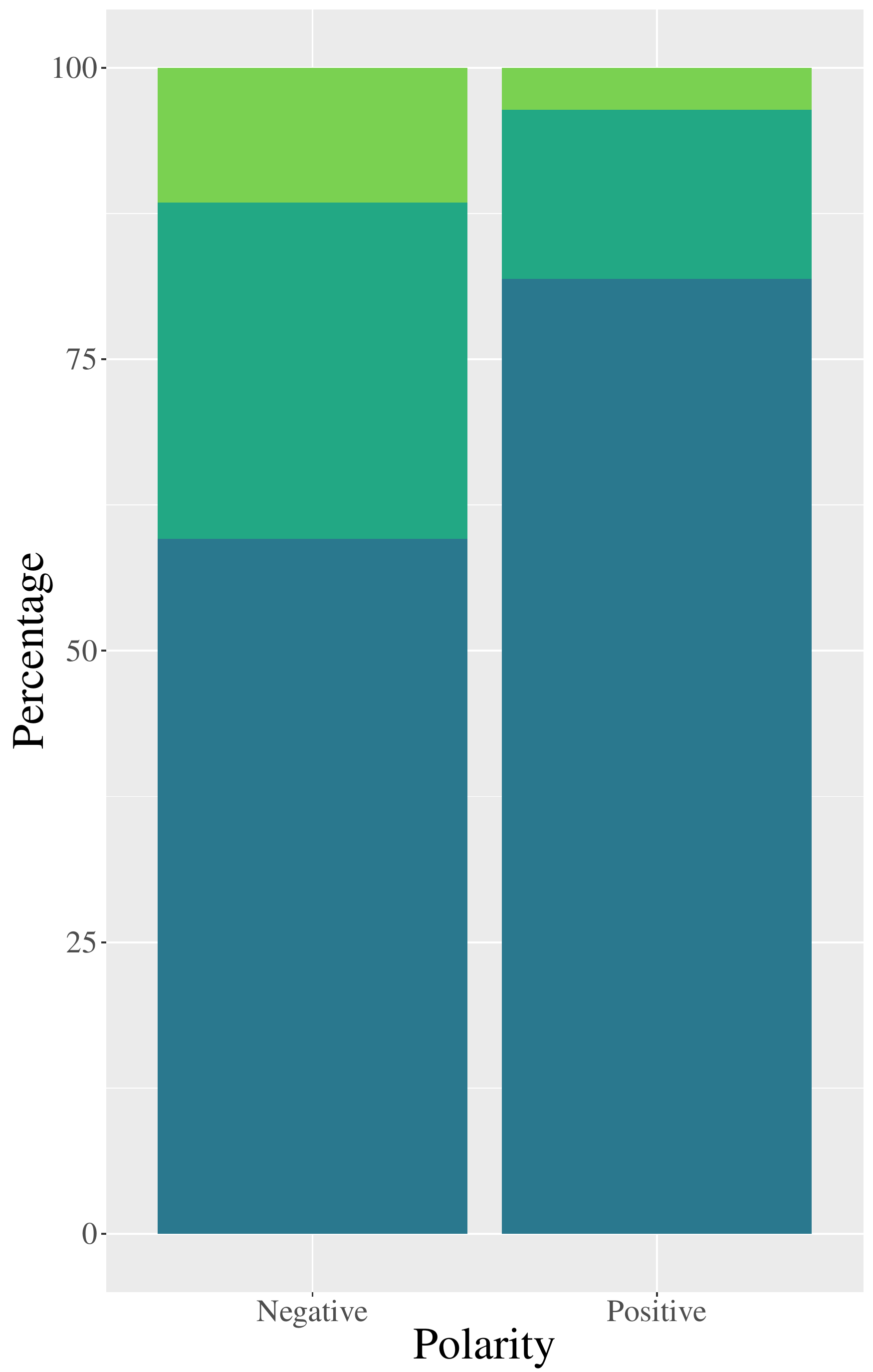}
        }%
        \subfigure[Chinese]{%
           \label{fig:neg_zh}
           \includegraphics[width=0.24\textwidth]{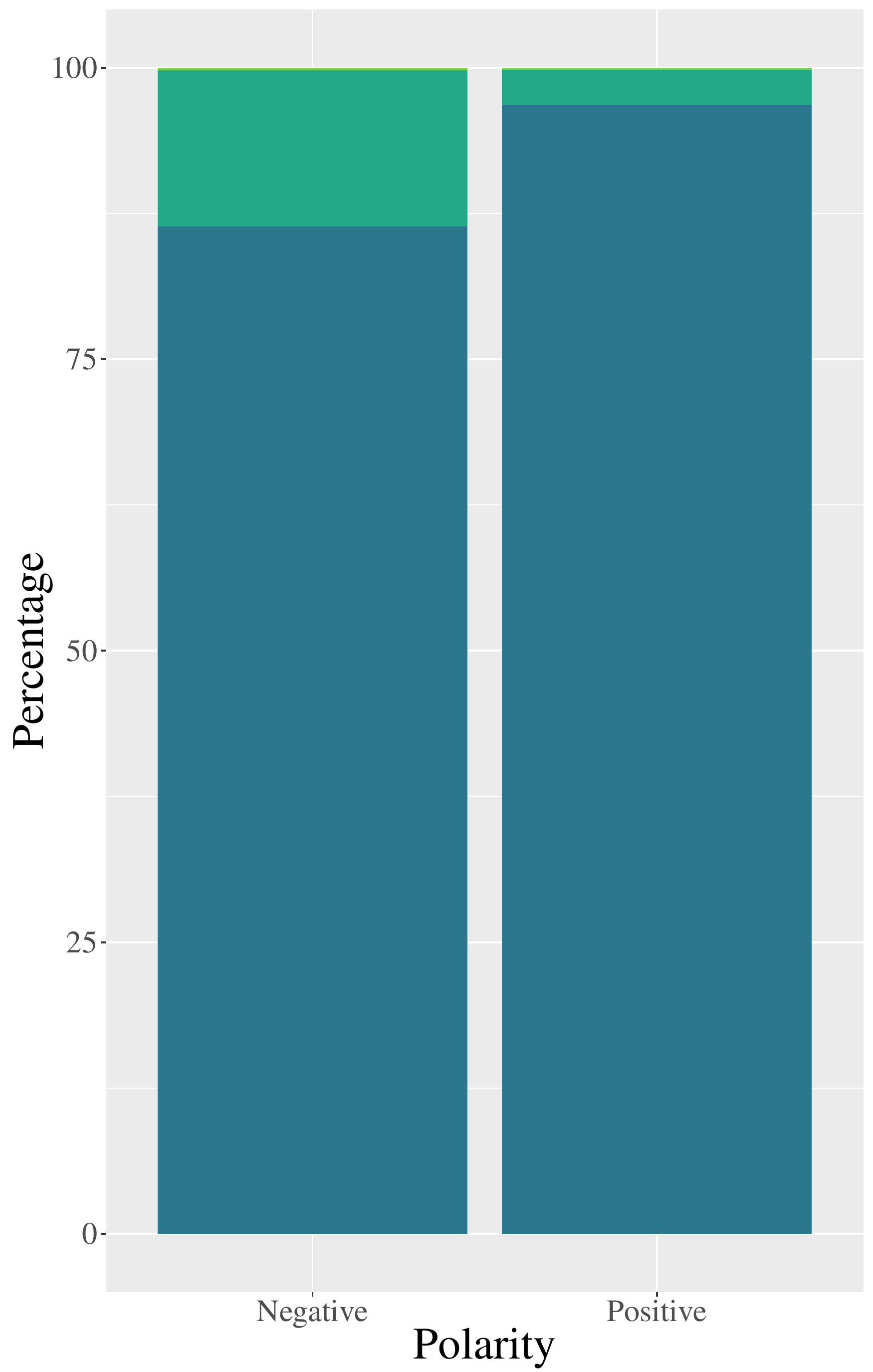}
        }%
        \subfigure[Dutch]{%
            \label{fig:neg_nl}
            \includegraphics[width=0.24\textwidth]{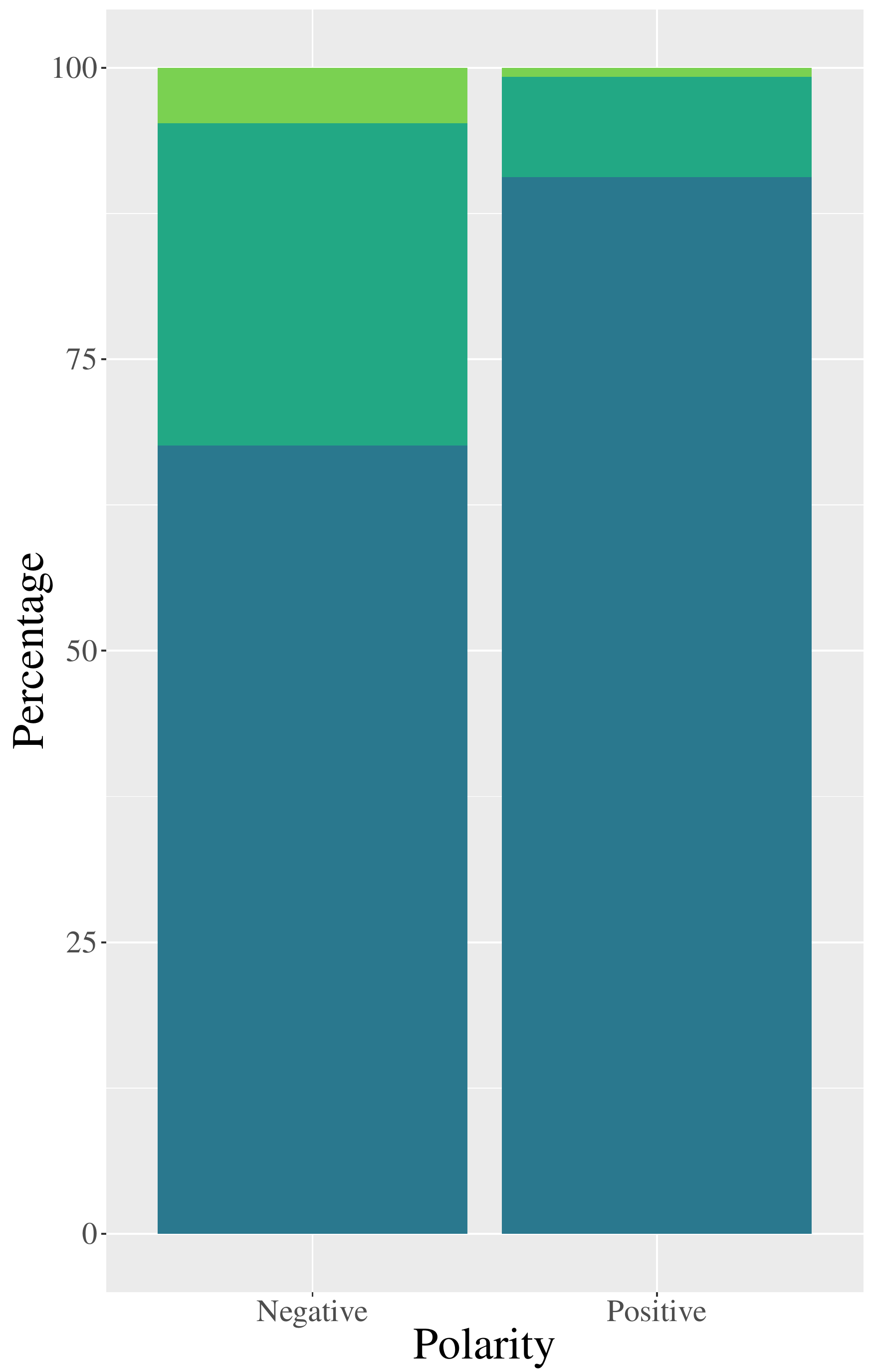}
        }%
        \subfigure[English]{%
           \label{fig:neg_en}
           \includegraphics[width=0.24\textwidth]{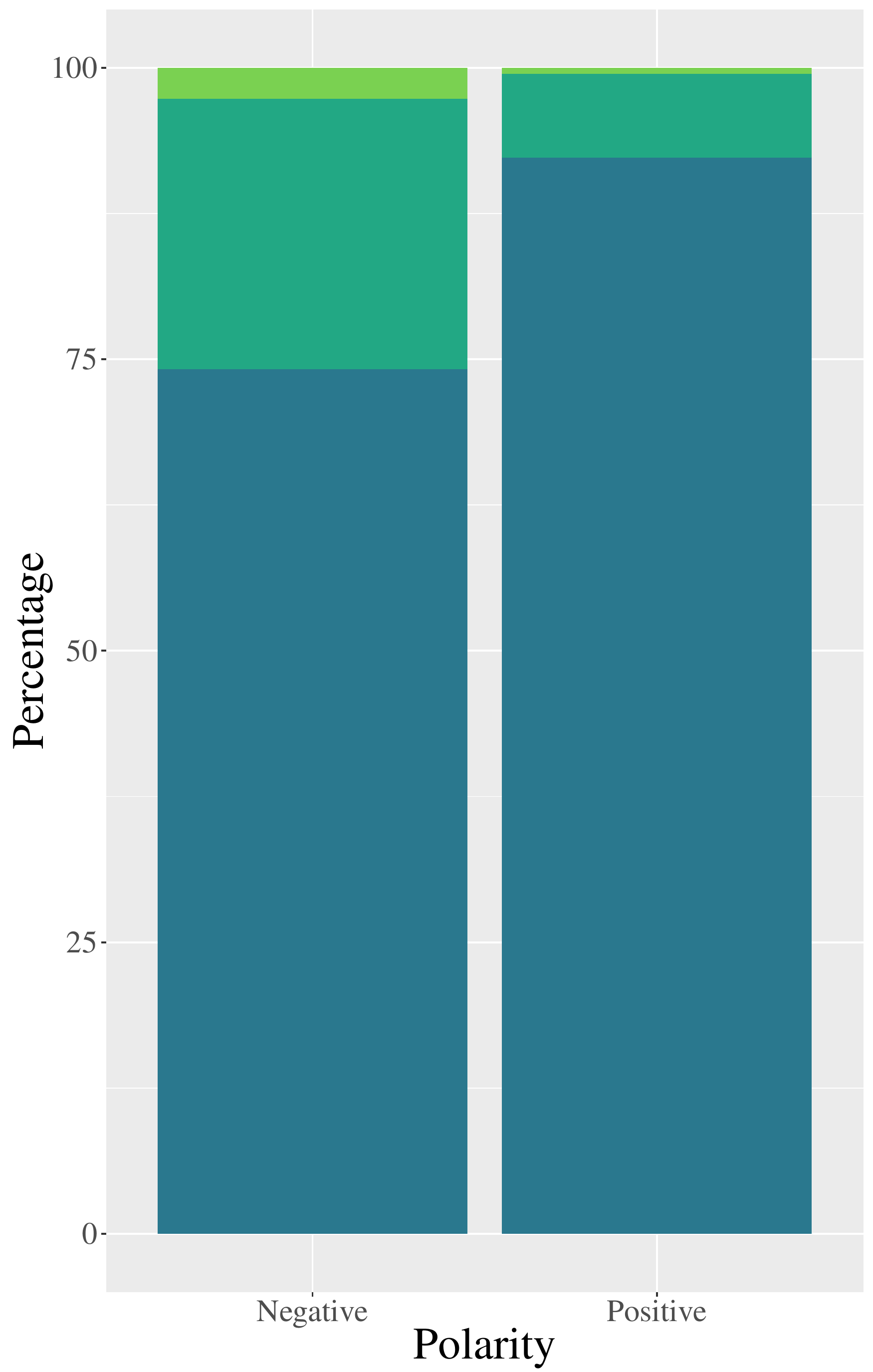}
        }%
        \\ 
        \subfigure[French]{%
            \label{fig:neg_fr}
            \includegraphics[width=0.24\textwidth]{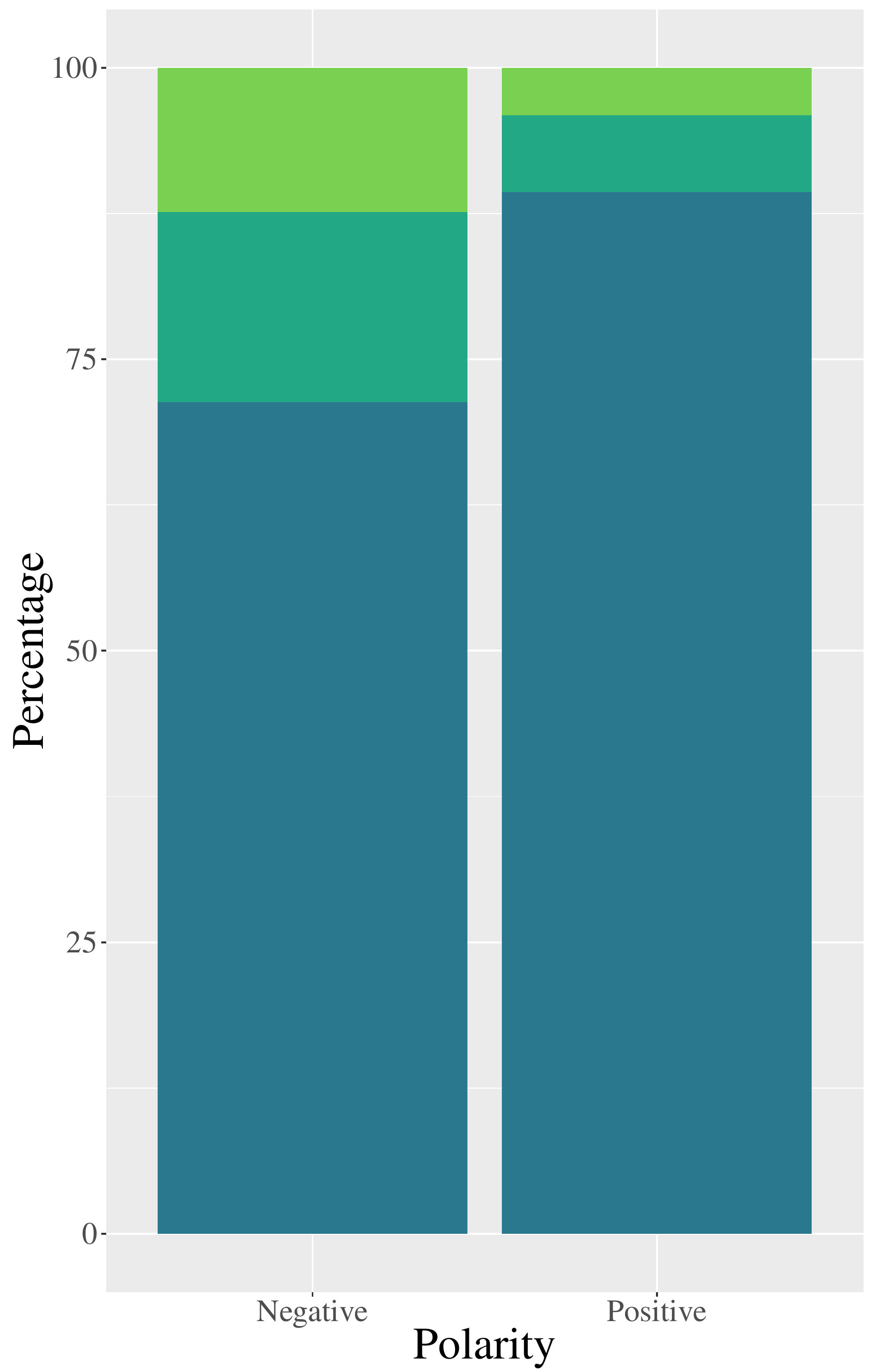}
        }%
        \subfigure[Italian]{%
           \label{fig:neg_it}
           \includegraphics[width=0.24\textwidth]{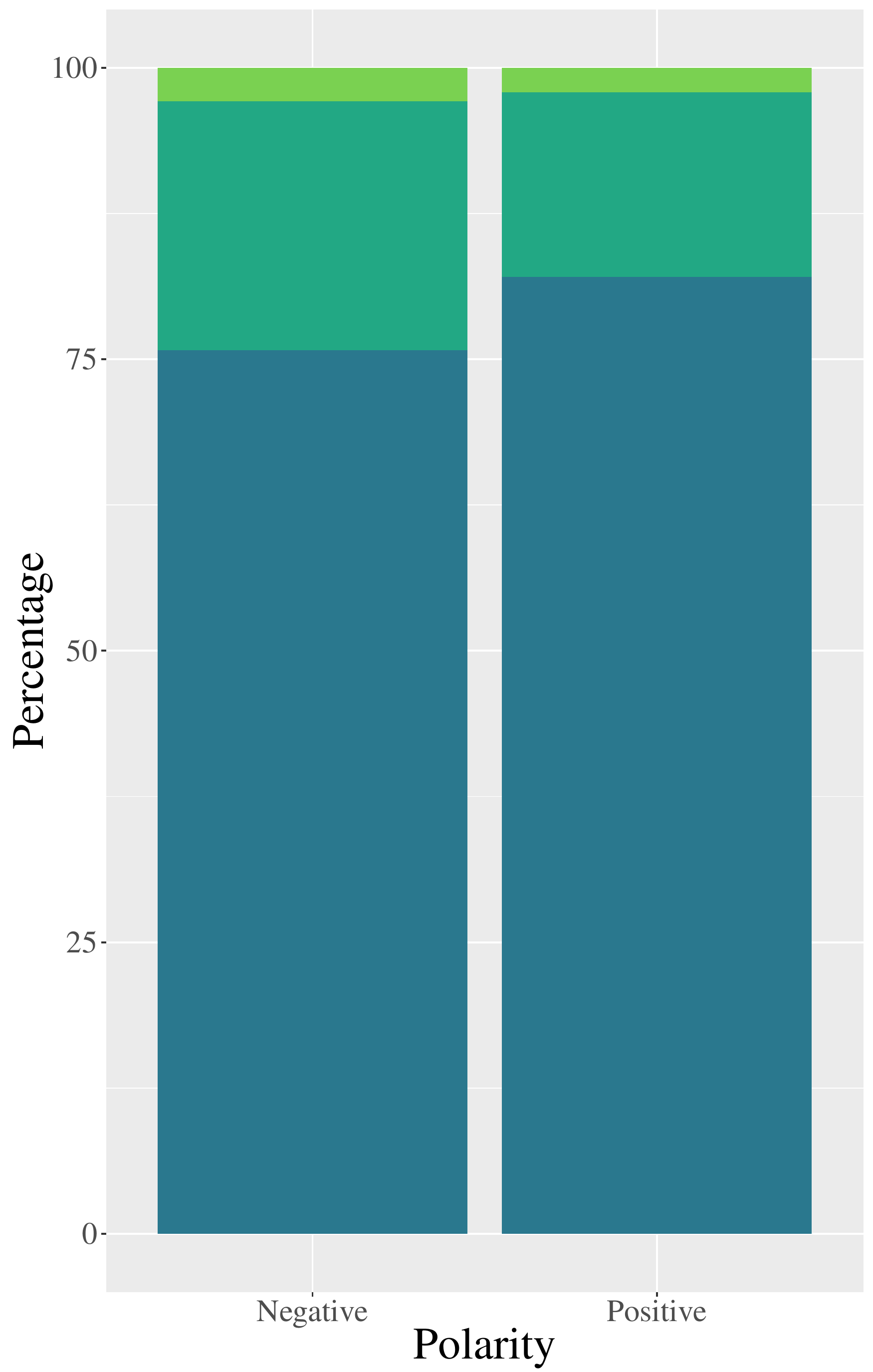}
        }%
        \subfigure[Russian]{%
            \label{fig:neg_ru}
            \includegraphics[width=0.24\textwidth]{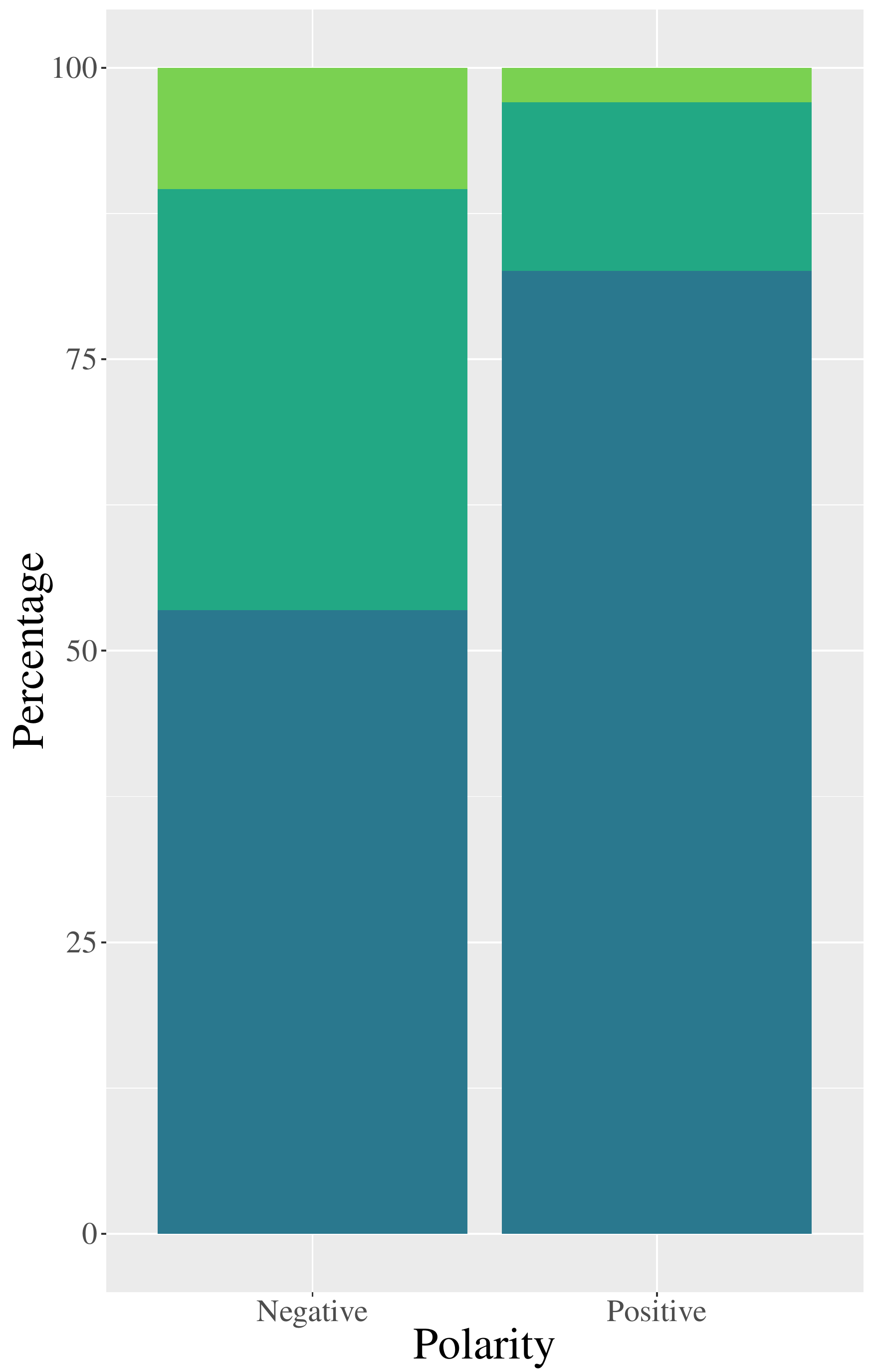}
        }%
        \subfigure[Spanish]{%
           \label{fig:neg_es}
           \includegraphics[width=0.24\textwidth]{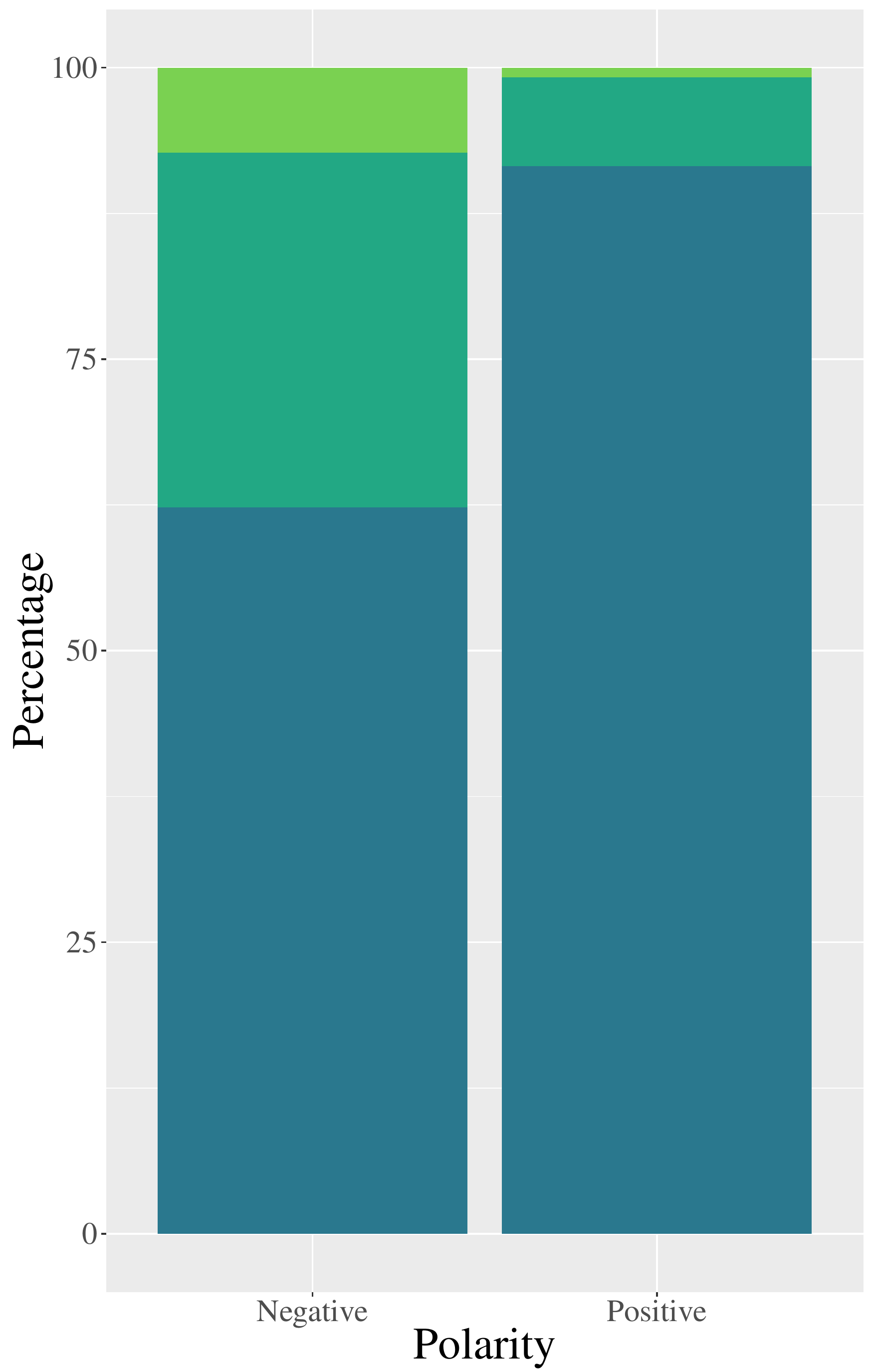}
        }%
        \\ 
    \end{center}
    \caption{%
        Percentages of negative (left) and positive (right) sentences with the same amount of negative grammatical markers. A count of 0 is represented in dark blue, 1 in light blue, and 2 or more in green.
     }%
   \label{fig:negcounts}
\end{figure*}

\section{Experimental Setup}
\label{sec:setup}
Now, we evaluate the quality of the distributed sentence representations from \S\ \ref{sec:algorithms} on Sentiment Analysis.  In \S\ \ref{sec:datasets} we introduce the datasets of all the considered languages, and the evaluation protocol in \S\ \ref{sec:evaluation}. Finally, to provide a potential performance ceiling, we compare the obtained results with those of a deep, state-of-art classifier, outlined in \S\ \ref{sec:bilstm}.

\subsection{Datasets}
\label{sec:datasets}
The data for training and testing are sourced from the SemEval 2016: Task 5 \cite{pontiki2016orphee}. These datasets provide customer reviews in 8 languages labelled with Aspect-Based Sentiment, i.e.,\ opinions about specific entities or attributes rather than generic stances. The languages include Arabic (hotels domain), Chinese (electronics), Dutch (restaurants and electronics), English (restaurants and electronics), French, Russian, Spanish, and Turkish (restaurants all). We mapped the labels to an overall polarity class (positive or negative) by selecting the majority class among the aspect-based sentiment classes for a given sentence. Note that no general sentiment for the sentence was included in this pool. Moreover, we added data for Italian (tweets) from the \textsc{sentipolc} shared task in \textsc{evalita} 2016 \cite{barbieri2016overview}. We discarded neutral stances from the corpus, and retained only positive and negative ones. Table~\ref{tab:datasets} shows the final size of the dataset partitions and the Wikipedia dumps. In Figure~\ref{fig:negcounts}, we report the percentage of sentences with the same amount of negative grammatical markers (e.g. the word \textit{not} and the suffix \textit{n't} in English) based on their polarity class. We discuss the impact of the variation of these percentages on the results in \S\ \ref{sec:results}.

\begin{table}[t]
\centering
{\small
\begin{tabularx}{\linewidth}{X |X|XX}
\toprule
\textbf{Language} & \textbf{Wikipedia Dumps} & \textbf{Train} & \textbf{Test}  \\
\cmidrule(lr){1-4}
\textit{Arabic} & 3406732 &	4570 &	1163\\
\textit{Chinese} & 8067971 &	2593 &	1011 \\
\textit{Dutch} & 11860559 & 2169	&	683\\
\textit{English} & 30000002 &	3584 &	1102 \\
\textit{French}	& 26024881 &	1410 &	534\\
\textit{Italian} & 15338617 &	4588	& 512 \\
\textit{Russian} & 16671224 &	2555 &	835\\
\textit{Spanish} & 22328668 &	1553 &	646\\
\textit{Turkish} & 3622336 &	1008 &	121\\
\bottomrule
\end{tabularx}
}%
\caption{Size of the data partitions (\# sentences).}
\label{tab:datasets}
\vspace{-3.5mm}
\end{table}

\subsection{Evaluation Protocol}
\label{sec:evaluation}
After mapping each sentence in the dataset to its distributed representation, we fed them to a Multi-Layer Perceptron (MLP), trained to detect the sentence polarity. In the MLP, a logistic regression layer is stacked onto a 60-dimensional hidden layer with a hyperbolic tangent activation. The weights were initialised from the random \textit{xavier} distribution \newcite{glorot2010understanding}. The cross-entropy loss was normalised with the L2-norm of the weights scaled by $\lambda=10^{-3}$. The optimisation with gradient descent ran for 20 epochs with early stopping. Batch size was 10 and the learning rate $10^{-2}$.

\subsection{Comparison with State-of-Art Models}
\label{sec:bilstm}
In addition to unsupervised distributed sentence representations, we test a bi-directional Long Short-Term Memory neural network (bi-LSTM) on the same task. This is a benchmark to compare against results of deep state-of-art architectures. The choice is based on the competitive results of this algorithm and on its sensitivity to word order. 
The accuracy of this architecture is 45.7 for 5-class and 85.4 for 2-class Sentiment Analysis on the standard dataset of the Stanford Sentiment Treebank.

The importance of word order is evident from the architecture of the network. In a recurrent model, the word embedding of a word $w_t$ at time $t$ is combined with the hidden state $h_{t-1}$ from the previous time step. The process is iterated throughout the whole sequence of words of a sentence. This model can be extended to multiple layers. LSTM is a refinement associating each time epoch with an input, control and
memory gate, in order to filter out irrelevant information \cite{hochreiter1997long}. This model is bi-directional if it is split in two branches reading simultaneously the sentence in opposite directions  \cite{schuster1997bidirectional}. 

Contrary to the evaluation protocol sketched in \S\ \ref{sec:evaluation}, the bi-LSTM does not utilise unsupervised sentence representations. Rather, it is trained directly on the datasets from \S\ \ref{sec:datasets}.
The optimisation ran for 20 epochs, with a batch size of 20 and a learning rate of $5 \cdot 10^{-2}$. The 60-dimensional hidden layer had a dropout probability of 0.2. Crucially, the word embeddings were initialised with the Skip-Gram model described in \S\ \ref{sssec:skipgram}. Since performance tends to vary depending on the initialisation, this ensures a fair comparison.

\begin{figure*}[t]
\centering
\includegraphics[width=\textwidth]{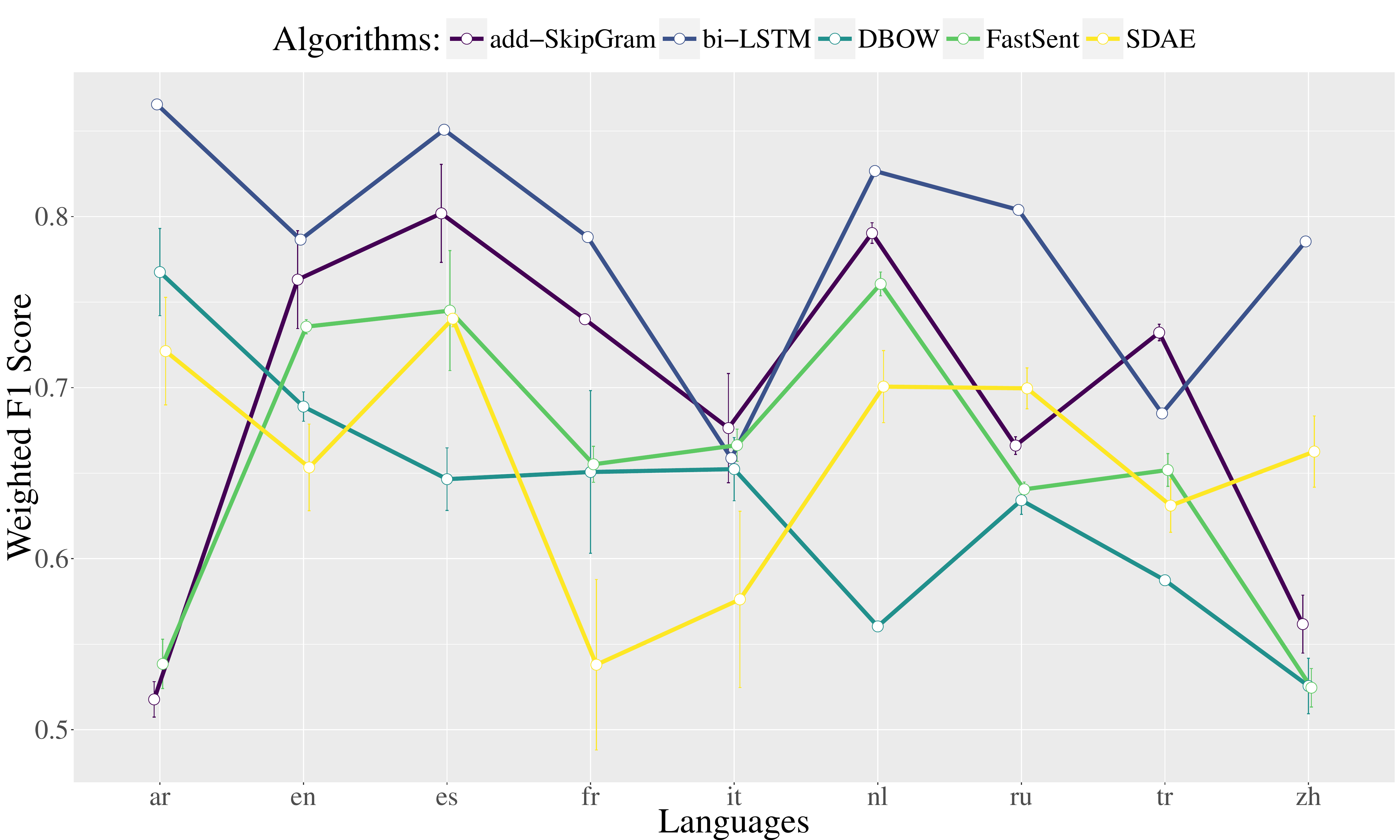}
\begin{tabular}{c|cccc|c}
& add-SkipGram & DBOW & FastSent & SDAE & bi-LSTM \\
\hline
ar & 51.76 $\pm$ 1.78 & \textbf{76.76} $\pm$ 4.42 & 53.85 $\pm$ 2.50 & \underline{72.13} $\pm$ 5.44 & 86.56\\
en & \textbf{76.31} $\pm$ 4.96 & 68.89 $\pm$ 1.49 & \underline{73.57} $\pm$ 0.71 & 65.33 $\pm$ 4.39 & 78.65\\
es & \textbf{80.19} $\pm$ 4.96 & 65.65 $\pm$ 3.17 & \underline{74.50} $\pm$ 6.07 & 74.03 $\pm$ 0.81 & 85.08\\
fr & \textbf{74.00} $\pm$ 0.43 & 65.07 $\pm$ 8.23 & \underline{65.52} $\pm$ 1.82 & 53.80 $\pm$ 8.63 & 78.80\\
it & \textbf{67.63} $\pm$ 5.52 & 65.24 $\pm$ 3.20 & \underline{66.63} $\pm$ 1.63 & 57.62 $\pm$ 8.92 & \textit{65.88}\\
nl & \textbf{79.04} $\pm$ 1.05 & 56.04 $\pm$ 0.00 & \underline{76.06} $\pm$ 1.20 & 70.06 $\pm$ 3.65 & 82.66\\
ru & \underline{66.61} $\pm$ 0.89 & 63.42 $\pm$ 1.43 & 64.05 $\pm$ 0.74 & \textbf{69.96} $\pm$ 2.07 & 80.39\\
tr & \textbf{73.22} $\pm$ 0.84 & 58.74 $\pm$ 0.00 & \underline{65.19} $\pm$ 1.66 & 63.10 $\pm$ 2.69 & \textit{68.49}\\
zh & \underline{56.17} $\pm$ 2.93 & 52.58 $\pm$ 2.80 & 52.46 $\pm$ 1.95 & \textbf{66.26} $\pm$ 3.61 & 78.55\\
\end{tabular}

\caption{Results of 5 different algorithms on 9 languages. Values report the mean Weighted F1 Score and the standard deviation. The best results per language are given in bold and the second-best is underlined. Data points where the ceiling is outperformed are in italics.}
\label{fig:results}
\end{figure*}


\section{Results}
\label{sec:results}
The results are displayed in Figure \ref{fig:results}. Weighted F1 scores were preferred over accuracy scores, since the two classes (positive and negative) are unbalanced. We decoded the unsupervised representations multiple times through different initialisation of the MLP weights, hence we report both the mean value and its standard deviation. The results are not straightforward: there is no algorithm outperforming the others in each language; unexpectedly not even the bi-LSTM used as a ceiling. However, the variation in performance follows certain trends, depending on the properties of languages and algorithms. We now examine: i) how performance is affected by the properties of the algorithms, such as those summarised in Table~\ref{tab:hyperparams}; ii) how typological features concerning negation and the text domain could make polarity harder to detect; iii) the interaction between negation and indefinite pronouns, by visualising the contribution of each word to the predicted class probabilities.

\subsection{Feature Sensitivity of the Algorithms}
The state-of-art bi-LSTM algorithm chosen as a ceiling is not the best choice in some languages (Italian, and Turkish). In these cases, it is always surpassed by the same model: additive Skip-Gram. The drop in Italian is possibly linked to its dataset in specific, since all the algorithms behave similarly badly. Turkish is possibly challenging for a recursive model because of the sparsity of its vocabulary. These cases, however, are not isolated: averaged word embeddings outperformed LSTMs in text similarity tasks \cite{arora2016simple} and provide a strong baseline in English \cite{Adi:2017iclr}.

In any case, the general high performance of additive Skip-Gram is noteworthy: it shows that a simple method achieves close-to-best results in almost every language among decoded distributed representations. This result is in line with other findings: \newcite{wieting2016towards} showed that word embeddings, once retrained and decoded by linear regression, beat many methods that generate sentence representations directly.

Moreover, the second-best method for languages is always FastSent, which is the only one hinging upon neighbouring sentences as features. This demonstrate that sentiment is encoded not only within a sentence, but also in its textual context. As a consequence, a relatively small and accessible dataset (Wikipedia) is sufficient to provide a reliable model in most languages. Nonetheless, the varying size of the dumps affects FastSent as well as the other unsupervised algorithms: limited data hinders them from learning faithful representations, as in Arabic, Chinese, and Turkish (see Table \ref{tab:datasets}).

In general, algorithms sensitive to the same features behave similarly, e.g. SDAE and bi-LSTM. They follow the same trend in relative improvements from one language to another. The generally low performance of SDAE could depend on the limited training time, which was necessary to evaluate the algorithm on the whole set of languages.




\begin{figure*}[th]
     \begin{center}
        \subfigure[Arabic positive]{%
            \label{fig:first}
            \includegraphics[width=0.24\textwidth]{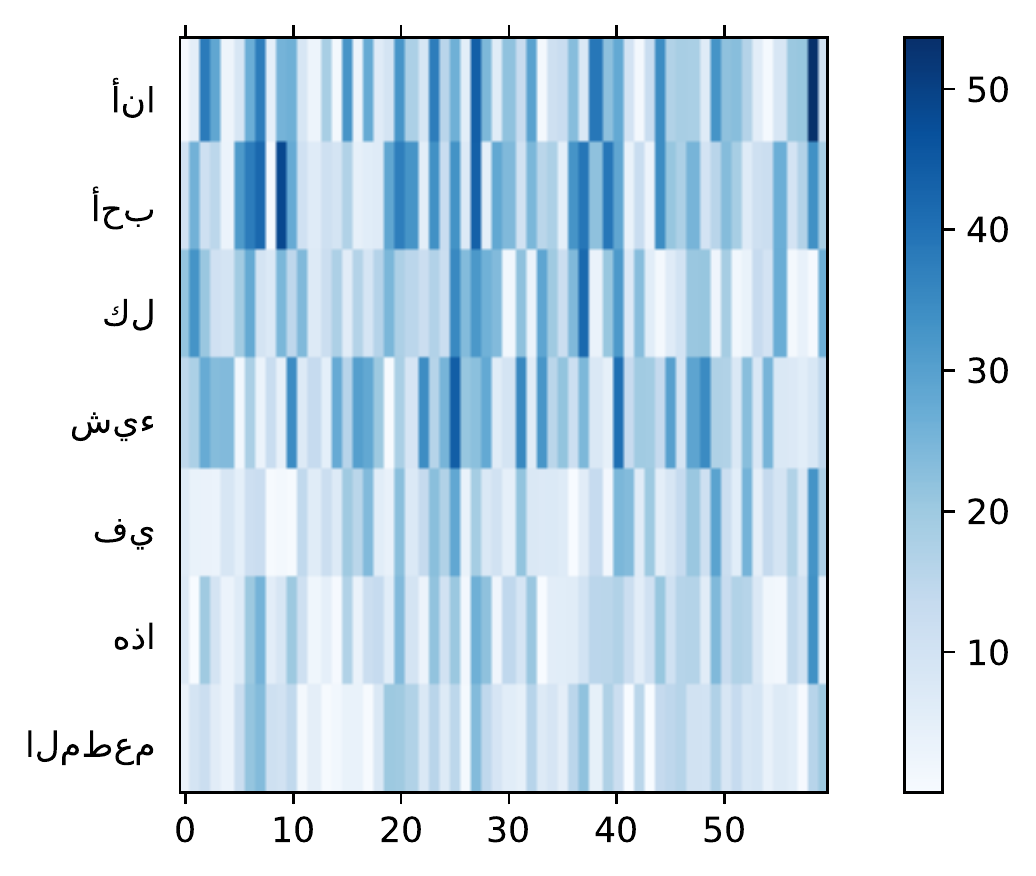}
        }%
        \subfigure[Arabic negative]{%
           \label{fig:second}
           \includegraphics[width=0.24\textwidth]{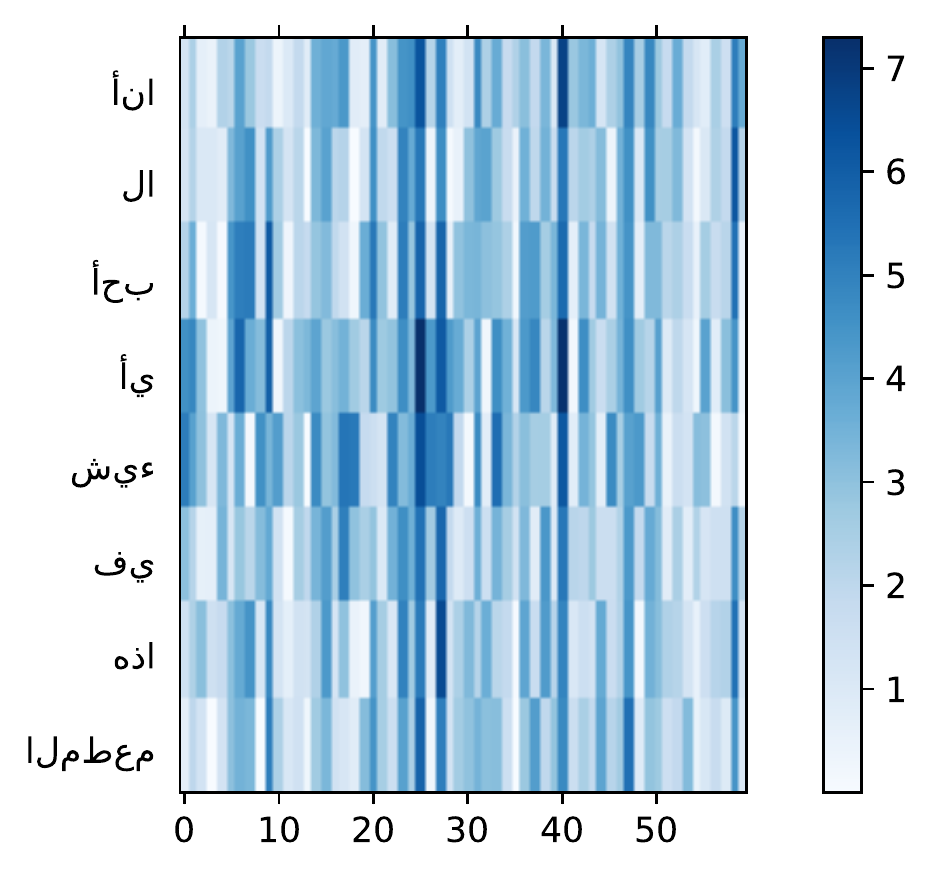}
        }%
        \subfigure[Spanish positive]{%
            \label{fig:fifth}
            \includegraphics[width=0.24\textwidth]{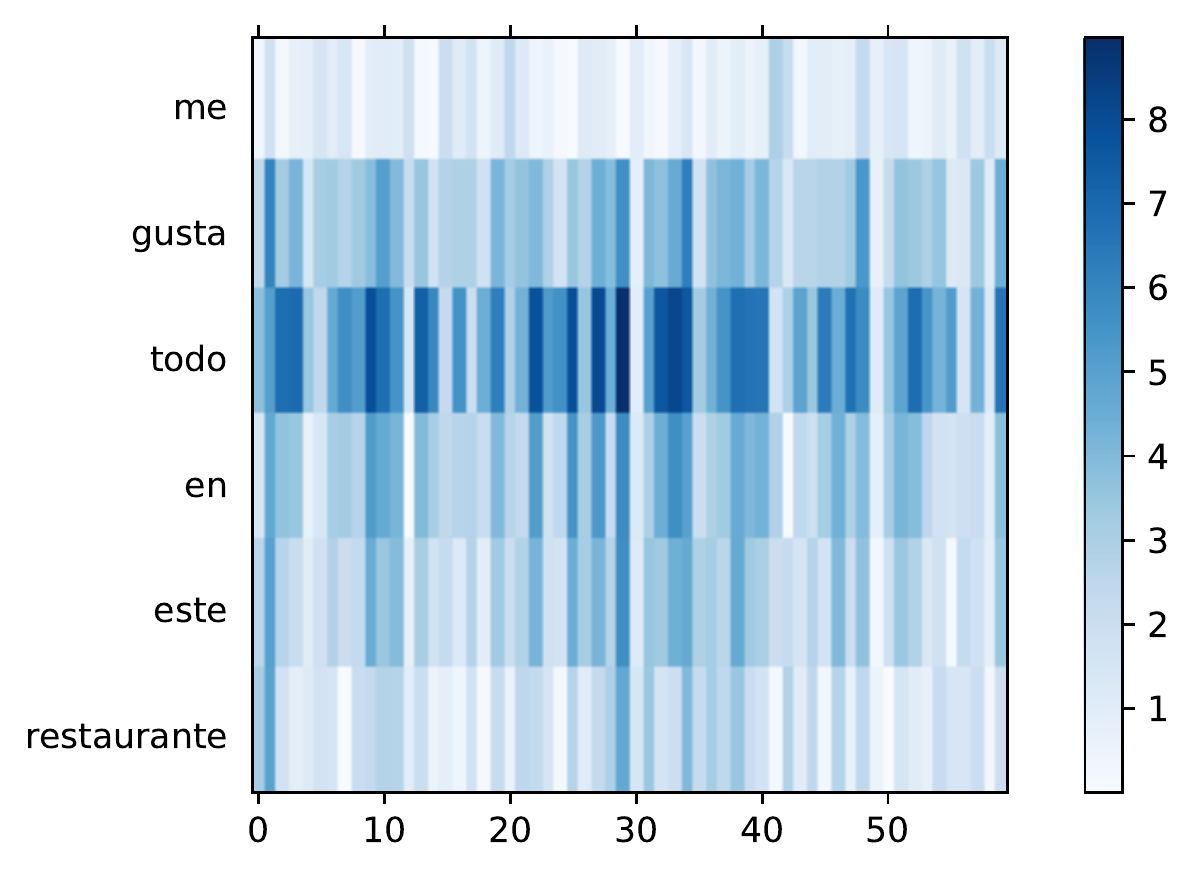}
        }%
        \subfigure[Spanish negative]{%
           \label{fig:sixth}
           \includegraphics[width=0.24\textwidth]{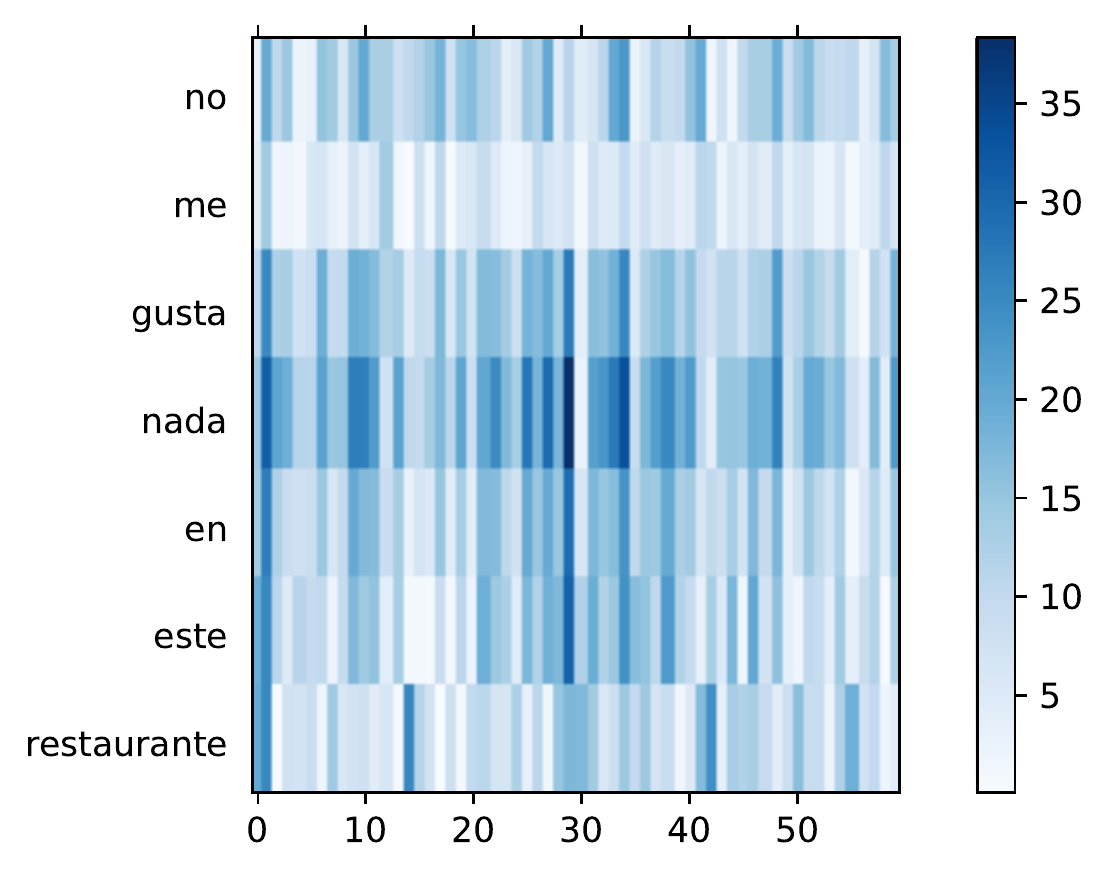}
        }
        \\ 
		\subfigure[Russian positive]{%
            \label{fig:13th}
            \includegraphics[width=0.24\textwidth]{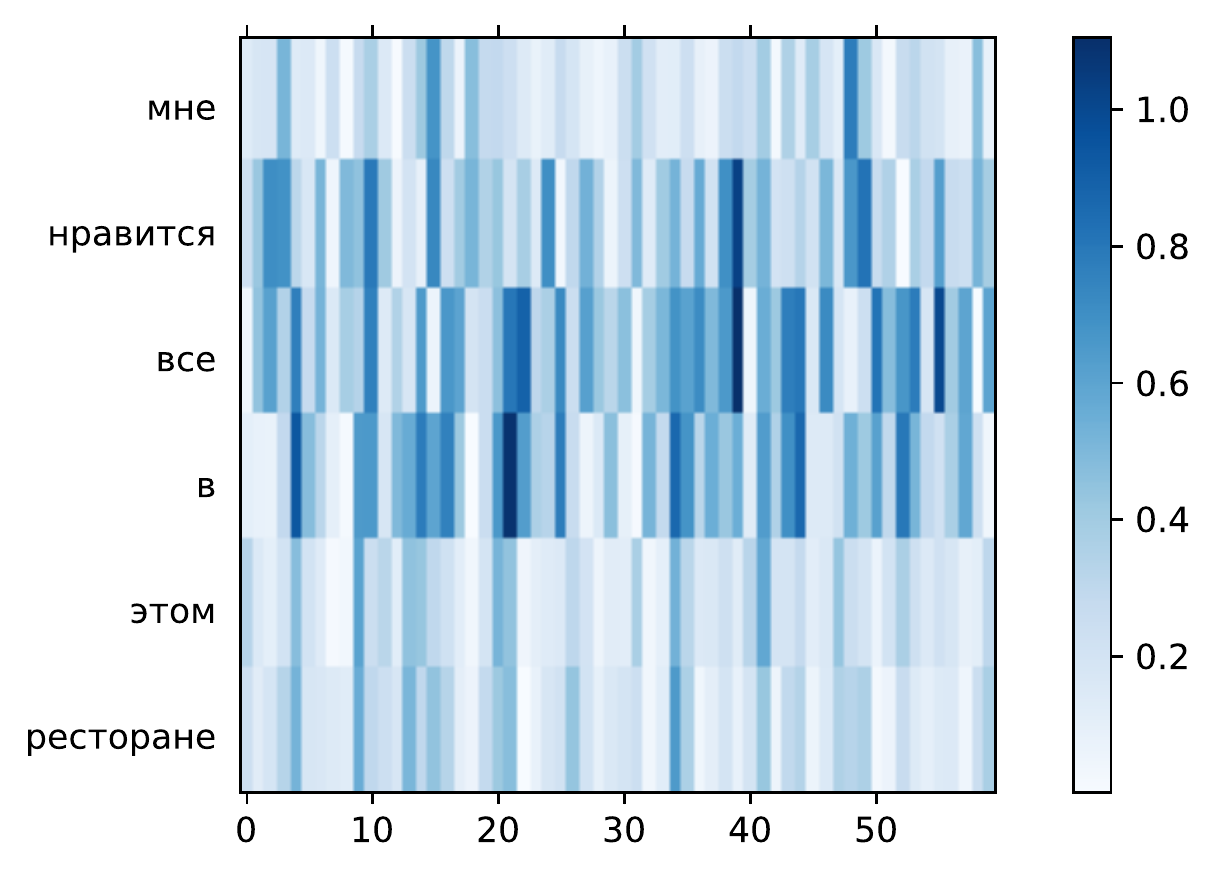}
        }%
        \subfigure[Russian negative]{%
           \label{fig:14th}
           \includegraphics[width=0.24\textwidth]{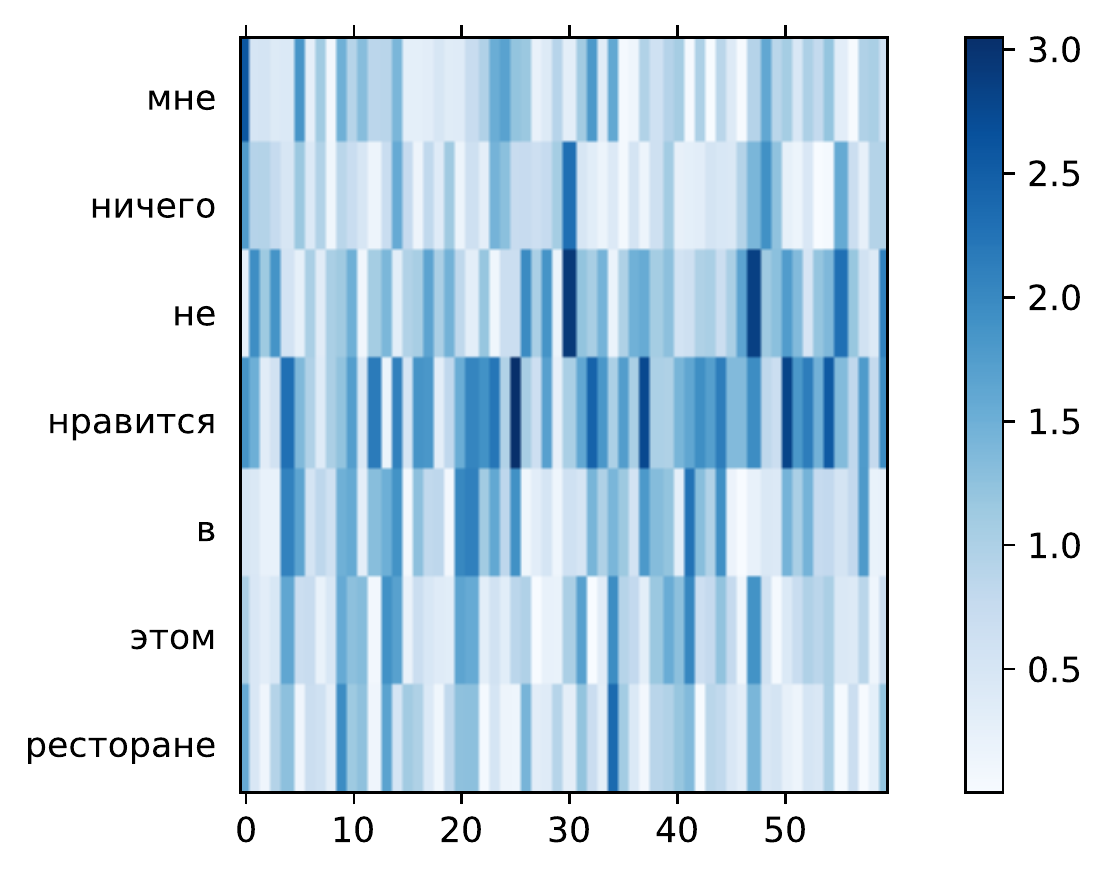}
        }%
\\ 
%
    \end{center}
    \caption{%
        Visualization of the derivative of the class scores with respect to the word embeddings.
     }%
   \label{fig:subfigures}
\end{figure*}

\subsection{Typology of Negation and Domain}
In some languages, the scores are very scattered: this fluctuation might be due to their peculiar morphological properties. In particular, Arabic is an introflexive language, Chinese is a radically isolating language, and Turkish an agglutinative language. On the other hand, the algorithms achieve better scores in the fusional languages, save Italian. 

A fine-grained analysis shows also that the performance is affected by the typology of the negation in each language, although negative markers appear in a reduced number of examples (see Figure \ref{fig:negcounts}). Semantically, negation is crucial in switching or mitigating the polarity of lexical items and phrases. 
Morpho-syntactically, negation is expressed through several constructions across the languages of the world. Constructions differ in many respects, which are classified as feature-value pairs in databases like the World Atlas of Language Structures \cite{wals}.\footnote{The features considered here for negation are 113A `Symmetric and Asymmetric Standard Negation', 114A `Subtypes of Asymmetric Standard Negation', 115A `Negative Indefinite Pronouns and Predicate Negation', and 143A `Order of Negative Morpheme and Verb'.}

Negation can affect the declarative verbal main clauses. In fact, negative clauses can be: i) symmetric, i.e.,\ identical to the affirmative counterpart except for the negative marker; ii) asymmetric, i.e.\ showing structural differences between negative and affirmative clauses (in constructions or paradigms); iii) showing mixed behaviour. Alterations concern for instance finiteness, the obligatory marking of unreality status, or the expression of verbal categories.
Secondly, negation interacts with indefinite pronoun (e.g.\ \textit{nobody}, \textit{nowhere}, \textit{never}). Negative indefinites can i) co-occur with standard negation; ii) be forbidden in concurrence; iii) display a mixed behaviour.
Finally, the relation of the negative marker with respect to verb is prone to change. Firstly, it can be either an affix or a prosodically independent word. Secondly, its position can be anchored to the verb (preceding, following, or both). Thirdly, negation can be omitted, doubled or even tripled. 
Performances seem to suffer the ambiguity in mapping between a negative marker and negative meaning. In fact, the bi-LSTM achieves lower scores in languages with asymmetric constructions (Chinese, English, and Turkish): the additional changes in the sentence construction and/or verb paradigm might create noise. Additional reasons of difficulty may occur when negation is doubled (French) or affixed (Turkish), since this makes negation redundant or sparse. On the other hand, add-SkipGram appears to be sensitive to the presence of negation: according to the counts in Figure~\ref{fig:negcounts}, when this is too pervasive (Arabic and Russian) or rare (Chinese), the scores tend to decrease.

These comments on the results based on linguistic properties can also suggest speculative solutions for future work. For algorithms based on sentence order, it is not clear whether the problem lies in the lack of wider collections of texts in some languages, or rather on the maximum amount of information about polarity that is learnt through a sentence-level distributional hypothesis. On the other hand, impairments of the other algorithms seem to be linked with redundancies and noise. Filtering out words that contribute to this effect might benefit the quality of the representation. Moreover, the sparsity due to cases where negation is an affix might be mitigated by introducing character-level features.


The other inherent source of variation is the text domain, on which the difficulty of the task depends \cite{glorot2011domain}. Although the unstructured nature of tweets could hinder the quality of the sentence representations in Italian, however, no clear effect is evident based on the other domains.

\subsection{Visualisation}
\label{sec:visualization}
Since languages vary in the ``polarity agreement'' between verbs and indefinite pronouns, algorithms may weigh these as features differently. We analyse their role through a visualizasion of the activation in the hidden layer of the bi-LSTM. In particular, we approximated the objective function through a linear function, and estimated the contribution of each word to the true class probability by computing the prime derivative of the output scores with respect to the embeddings. This technique is presented and detailed by \newcite{li2015visualizing}. The visualised hidden layers are shown in Figure~\ref{fig:subfigures}, whereas the sentences used as input are glossed in Ex.~\eqref{ex1} (Arabic), Ex.~\eqref{ex2} (Spanish), and Ex.~\ref{ex3} (Russian).

\begin{examples}
\item \gll `ana `uhibu kl shay` fi hadha almataeim / `ana la `uhibu `ayu shay` fi hadha almataeam
      \textsc{1sg} like.\textsc{npst.1sg} every thing in this restaurant \textit{/} \textsc{1sg} not.\textsc{npst} like.\textsc{npst.1sg} any thing in this restaurant
      \glt 
      \glend
      \label{ex1}
\item \gll me gust-a todo en est-e restaurant-e / no me gust-a nada en est-e restaurant-e
      \textsc{1sg.dat} like-\textsc{3sg} everything in this\textsc{-sg} restaurant\textsc{-sg} \textit{/} not \textsc{1sg.dat} like-\textsc{3sg} nothing in this\textsc{-sg} restaurant\textsc{-sg}
      \glt 
      \glend
      \label{ex2}
\item \gll mne nr\'av-itsja vs-jo v \'et-om restor\'ane / mne ni-\v{c}ev\'o ne nr\'av-itsja v \'et-om restor\'an-e
      \textsc{1sg.dat} like\textsc{.impv}-\textsc{prs.3sg} all\textsc{-nom.sg} in this\textsc{-prep.sg} restaurant\textsc{-prep.sg} \textit{/} \textsc{1sg.dat} nothing\textsc{-gen} not like\textsc{.impv}-\textsc{prs.3sg} in this\textsc{-prep.sg} restaurant\textsc{-prep.sg}
      \glt 
      \glend
      \label{ex3}
\end{examples}
\vspace{-12pt}

The two compared sentences correspond to the translation of two English sentences. The first is positive: `I like everything in this restaurant'; the second is negative: `I don't like anything in this restaurant'. These include a domain-specific but sentiment-neutral word that plays the role of a touchstone. The more a cell tends to blue, the higher its activation. In some languages (e.g.\ Arabic), the sentiment verb elicits a stronger reaction in the positive polarity, whereas the indefinite pronoun dominates in the negative polarity. In several other languages (e.g.\ Spanish), indefinite pronouns are more relevant than any other feature. In Russian, only sentiment verbs always provoke a reaction. These differences might be related to the ``polarity agreement'' of these languages, which happens always, sometimes, and never, respectively. In some other languages, however, no evidence is found of any similar activation pattern.

\section{Conclusion}
In this work, we examined how much sentiment polarity information is retained by distributed representations of sentences in multiple typologically diverse languages. We generated the representations through various algorithms, sensitive to different properties from training corpora (e.g, word or sentence order). We decoded them through a simple MLP and compared their performance with one of the state-of-art algorithms for Sentiment Analysis: bi-directional LSTM. Unexpectedly, for some languages the bi-directional LSTM is outperformed by unsupervised strategies like the addition of the word embeddings obtained from a Skip-Gram model. This model, in turn, surpasses more sophisticated algorithms for most of the languages. This demonstrates i) that no algorithm is the best across the board; and ii) that some simple models are to be preferred even for downstream tasks, which partially contrasts with the conclusions of \newcite{hill2016learning}. Moreover, representation algorithms sensitive to word order have similar trends, but they do not always achieve performance superior to algorithms based on the sentence order. Finally, some properties of languages (i.e.\ their type of negation) appear to have an impact on the scores: in particular, the asymmetry of negative and affirmative clauses and the doubling of negative markers.

\section*{Acknowledgements}
This work was supported by the ERC Consolidator Grant LEXICAL (648909). The authors would like to thank the anonymous reviewers for their helpful suggestions and comments.

\bibliographystyle{acl_natbib}
\bibliography{acl2017_refs}

\begin{thebibliography}{}
\expandafter\ifx\csname natexlab\endcsname\relax\def\natexlab#1{#1}\fi

\bibitem[{Adi et~al.(2017)Adi, Kermany, Belinkov, Lavi, and
  Goldberg}]{Adi:2017iclr}
Yossi Adi, Einat Kermany, Yonatan Belinkov, Ofer Lavi, and Yoav Goldberg. 2017.
\newblock \href{http://arxiv.org/abs/1608.04207}{Fine-grained analysis of
  sentence embeddings using auxiliary prediction tasks}.
\newblock In {\em Proceedings of ICLR\/}.
\newblock
  \href{http://arxiv.org/abs/1608.04207}{http://arxiv.org/abs/1608.04207}.

\bibitem[{Almeida et~al.(2015)Almeida, Pinto, Figueira, Mendes, and
  Martins}]{almeida2015aligning}
Mariana~SC Almeida, Cl{\'a}udia Pinto, Helena Figueira, Pedro Mendes, and
  Andr{\'e}~FT Martins. 2015.
\newblock Aligning opinions: Cross-lingual opinion mining with dependencies.
\newblock In {\em Proc. of the Annual Meeting of the Association for
  Computational Linguistics\/}.

\bibitem[{Arora et~al.(2016)Arora, Liang, and Ma}]{arora2016simple}
Sanjeev Arora, Yingyu Liang, and Tengyu Ma. 2016.
\newblock A simple but tough-to-beat baseline for sentence embeddings.
\newblock In {\em ICLR 2017\/}.

\bibitem[{Balahur and Turchi(2014)}]{balahur2014comparative}
Alexandra Balahur and Marco Turchi. 2014.
\newblock Comparative experiments using supervised learning and machine
  translation for multilingual sentiment analysis.
\newblock {\em Computer Speech \& Language\/} 28(1):56--75.

\bibitem[{Barbieri et~al.(2016)Barbieri, Basile, Croce, Nissim, Novielli, and
  Patti}]{barbieri2016overview}
Francesco Barbieri, Valerio Basile, Danilo Croce, Malvina Nissim, Nicole
  Novielli, and Viviana Patti. 2016.
\newblock Overview of the evalita 2016 sentiment polarity classification task.
\newblock In {\em Proceedings of Third Italian Conference on Computational
  Linguistics (CLiC-it 2016) \& Fifth Evaluation Campaign of Natural Language
  Processing and Speech Tools for Italian. Final Workshop (EVALITA 2016)\/}.

\bibitem[{Boyd-Graber and Resnik(2010)}]{boyd2010holistic}
Jordan Boyd-Graber and Philip Resnik. 2010.
\newblock Holistic sentiment analysis across languages: Multilingual supervised
  latent dirichlet allocation.
\newblock In {\em Proceedings of the 2010 Conference on Empirical Methods in
  Natural Language Processing\/}. Association for Computational Linguistics,
  pages 45--55.

\bibitem[{Chen et~al.(2016)Chen, Athiwaratkun, Sun, Weinberger, and
  Cardie}]{chen2016adversarial}
Xilun Chen, Ben Athiwaratkun, Yu~Sun, Kilian Weinberger, and Claire Cardie.
  2016.
\newblock Adversarial deep averaging networks for cross-lingual sentiment
  classification.
\newblock {\em arXiv preprint arXiv:1606.01614\/} .

\bibitem[{Conneau et~al.(2017)Conneau, Kiela, Schwenk, Barrault, and
  Bordes}]{Conneau:2017arxiv}
Alexis Conneau, Douwe Kiela, Holger Schwenk, Lo{\"{\i}}c Barrault, and Antoine
  Bordes. 2017.
\newblock \href{http://arxiv.org/abs/1705.02364}{Supervised learning of
  universal sentence representations from natural language inference data}.
\newblock {\em CoRR\/} abs/1705.02364.
\newblock
  \href{http://arxiv.org/abs/1705.02364}{http://arxiv.org/abs/1705.02364}.

\bibitem[{Dahl(1979)}]{dahl1979typology}
{\"O}sten Dahl. 1979.
\newblock Typology of sentence negation.
\newblock {\em Linguistics\/} 17(1-2):79--106.

\bibitem[{Denecke(2008)}]{denecke2008using}
Kerstin Denecke. 2008.
\newblock Using sentiwordnet for multilingual sentiment analysis.
\newblock In {\em Data Engineering Workshop, 2008. ICDEW 2008. IEEE 24th
  International Conference on\/}. pages 507--512.

\bibitem[{Dryer and Haspelmath(2013)}]{wals}
Matthew~S. Dryer and Martin Haspelmath, editors. 2013.
\newblock {\em WALS Online\/}.
\newblock Max Planck Institute for Evolutionary Anthropology, Leipzig.
\newblock \href{http://wals.info/}{http://wals.info/}.

\bibitem[{Fern{\'a}ndez et~al.(2015)Fern{\'a}ndez, Esuli, and
  Sebastiani}]{fernandez2015distributional}
Alejandro~Moreo Fern{\'a}ndez, Andrea Esuli, and Fabrizio Sebastiani. 2015.
\newblock Distributional correspondence indexing for cross-lingual and
  cross-domain sentiment classification.
\newblock {\em Journal of Artificial Intelligence Research\/} 55:131--163.

\bibitem[{Glorot and Bengio(2010)}]{glorot2010understanding}
Xavier Glorot and Yoshua Bengio. 2010.
\newblock Understanding the difficulty of training deep feedforward neural
  networks.
\newblock In {\em Aistats\/}. volume~9, pages 249--256.

\bibitem[{Glorot et~al.(2011)Glorot, Bordes, and Bengio}]{glorot2011domain}
Xavier Glorot, Antoine Bordes, and Yoshua Bengio. 2011.
\newblock Domain adaptation for large-scale sentiment classification: A deep
  learning approach.
\newblock In {\em Proceedings of the 28th international conference on machine
  learning (ICML-11)\/}. pages 513--520.

\bibitem[{Hill et~al.(2016)Hill, Cho, and Korhonen}]{hill2016learning}
Felix Hill, Kyunghyun Cho, and Anna Korhonen. 2016.
\newblock Learning distributed representations of sentences from unlabelled
  data.
\newblock {\em arXiv preprint arXiv:1602.03483\/} .

\bibitem[{Hochreiter and Schmidhuber(1997)}]{hochreiter1997long}
Sepp Hochreiter and J{\"u}rgen Schmidhuber. 1997.
\newblock Long short-term memory.
\newblock {\em Neural computation\/} 9(8):1735--1780.

\bibitem[{Iyyer et~al.(2015)Iyyer, Manjunatha, Boyd-Graber, and
  Daum{\'e}~III}]{iyyer2015deep}
Mohit Iyyer, Varun Manjunatha, Jordan~L Boyd-Graber, and Hal Daum{\'e}~III.
  2015.
\newblock Deep unordered composition rivals syntactic methods for text
  classification.
\newblock In {\em ACL (1)\/}. pages 1681--1691.

\bibitem[{Kiros et~al.(2015)Kiros, Zhu, Salakhutdinov, Zemel, Urtasun,
  Torralba, and Fidler}]{kiros2015skip}
Ryan Kiros, Yukun Zhu, Ruslan~R Salakhutdinov, Richard Zemel, Raquel Urtasun,
  Antonio Torralba, and Sanja Fidler. 2015.
\newblock Skip-thought vectors.
\newblock In {\em Advances in neural information processing systems\/}. pages
  3294--3302.

\bibitem[{Kokkinos and Potamianos(2017)}]{kokkinos2017structural}
Filippos Kokkinos and Alexandros Potamianos. 2017.
\newblock Structural attention neural networks for improved sentiment analysis.
\newblock In {\em EACL 2017\/}.

\bibitem[{Le and Mikolov(2014)}]{le2014distributed}
Quoc~V Le and Tomas Mikolov. 2014.
\newblock Distributed representations of sentences and documents.
\newblock In {\em ICML\/}. volume~14, pages 1188--1196.

\bibitem[{Li et~al.(2015)Li, Chen, Hovy, and Jurafsky}]{li2015visualizing}
Jiwei Li, Xinlei Chen, Eduard Hovy, and Dan Jurafsky. 2015.
\newblock Visualizing and understanding neural models in nlp.
\newblock {\em arXiv preprint arXiv:1506.01066\/} .

\bibitem[{Lu et~al.(2011)Lu, Tan, Cardie, and Tsou}]{lu2011joint}
Bin Lu, Chenhao Tan, Claire Cardie, and Benjamin~K Tsou. 2011.
\newblock Joint bilingual sentiment classification with unlabeled parallel
  corpora.
\newblock In {\em Proceedings of the 49th Annual Meeting of the Association for
  Computational Linguistics: Human Language Technologies-Volume 1\/}.
  Association for Computational Linguistics, pages 320--330.

\bibitem[{Marelli et~al.(2014)Marelli, Bentivogli, Baroni, Bernardi, Menini,
  and Zamparelli}]{marelli2014semeval}
Marco Marelli, Luisa Bentivogli, Marco Baroni, Raffaella Bernardi, Stefano
  Menini, and Roberto Zamparelli. 2014.
\newblock Semeval-2014 task 1: Evaluation of compositional distributional
  semantic models on full sentences through semantic relatedness and textual
  entailment.
\newblock {\em SemEval-2014\/} .

\bibitem[{Mikolov et~al.(2013)Mikolov, Sutskever, Chen, Corrado, and
  Dean}]{mikolov2013distributed}
Tomas Mikolov, Ilya Sutskever, Kai Chen, Greg~S Corrado, and Jeff Dean. 2013.
\newblock Distributed representations of words and phrases and their
  compositionality.
\newblock In {\em Advances in neural information processing systems\/}. pages
  3111--3119.

\bibitem[{Milajevs et~al.(2014)Milajevs, Kartsaklis, Sadrzadeh, and
  Purver}]{milajevs10evaluating}
Dmitrijs Milajevs, Dimitri Kartsaklis, Mehrnoosh Sadrzadeh, and Matthew Purver.
  2014.
\newblock Evaluating neural word representations in tensor-based compositional
  settings.
\newblock {\em idea\/} 10(47):39.

\bibitem[{Mitchell and Lapata(2010)}]{mitchell2010composition}
Jeff Mitchell and Mirella Lapata. 2010.
\newblock Composition in distributional models of semantics.
\newblock {\em Cognitive science\/} 34(8):1388--1429.

\bibitem[{Mousa and Schuller(2017)}]{mousacontextual}
Amr El-Desoky Mousa and Bj{\"o}rn Schuller. 2017.
\newblock Contextual bidirectional long short-term memory recurrent neural
  network language models: A generative approach to sentiment analysis.
\newblock In {\em EACL 2017\/}.

\bibitem[{Polajnar et~al.(2015)Polajnar, Rimell, and
  Clark}]{polajnar2015exploration}
Tamara Polajnar, Laura Rimell, and Stephen Clark. 2015.
\newblock An exploration of discourse-based sentence spaces for compositional
  distributional semantics.
\newblock In {\em Workshop on Linking Models of Lexical, Sentential and
  Discourse-level Semantics (LSDSem)\/}. page~1.

\bibitem[{Poliak et~al.(2017)Poliak, Rastogi, Martin, and
  Van~Durme}]{poliak2017efficient}
Adam Poliak, Pushpendre Rastogi, M~Patrick Martin, and Benjamin Van~Durme.
  2017.
\newblock Efficient, compositional, order-sensitive n-gram embeddings.
\newblock {\em EACL 2017\/} page 503.

\bibitem[{Ponti(2016)}]{ponti2016divergence}
Edoardo~Maria Ponti. 2016.
\newblock Divergence from syntax to linear order in ancient greek lexical
  networks.
\newblock In {\em The 29th International {FLAIRS} Conference\/}.

\bibitem[{Pontiki et~al.(2016)Pontiki, Galanis, Papageorgiou, Androutsopoulos,
  Manandhar, Al-Smadi, Al-Ayyoub, Zhao, Qin, De~Clercq, Hoste, Apidianaki,
  Tannier, Loukachevitch, Kotelnikov, Bel, Jim{\'e}nez-Zafra, , and
  Eryigit}]{pontiki2016orphee}
Maria Pontiki, Dimitrios Galanis, Haris Papageorgiou, Ion Androutsopoulos,
  Suresh Manandhar, Mohammad Al-Smadi, Mahmoud Al-Ayyoub, Yanyan Zhao, Bing
  Qin, Orph{\'e}e De~Clercq, V{\'e}ronique Hoste, Marianna Apidianaki, Xavier
  Tannier, Natalia Loukachevitch, Evgeny Kotelnikov, Nuria Bel, Salud~Mar{\i}a
  Jim{\'e}nez-Zafra, , and G{\"u}lsen Eryigit. 2016.
\newblock Semeval-2016 task 5: Aspect based sentiment analysis.
\newblock In {\em Proceedings of the 10th International Workshop on Semantic
  Evaluation, SemEval\/}. volume~16.

\bibitem[{Rimell et~al.(2016)Rimell, Maillard, Polajnar, and
  Clark}]{rimell2016relpron}
Laura Rimell, Jean Maillard, Tamara Polajnar, and Stephen Clark. 2016.
\newblock Relpron: A relative clause evaluation dataset for compositional
  distributional semantics.
\newblock {\em Computational Linguistics\/} .

\bibitem[{Schuster and Paliwal(1997)}]{schuster1997bidirectional}
Mike Schuster and Kuldip~K Paliwal. 1997.
\newblock Bidirectional recurrent neural networks.
\newblock {\em IEEE Transactions on Signal Processing\/} 45(11):2673--2681.

\bibitem[{Socher et~al.(2013)Socher, Perelygin, Wu, Chuang, Manning, Ng, Potts
  et~al.}]{socher2013recursive}
Richard Socher, Alex Perelygin, Jean~Y Wu, Jason Chuang, Christopher~D Manning,
  Andrew~Y Ng, Christopher Potts, et~al. 2013.
\newblock Recursive deep models for semantic compositionality over a sentiment
  treebank.
\newblock In {\em Proceedings of the conference on empirical methods in natural
  language processing (EMNLP)\/}. Citeseer, volume 1631, page 1642.

\bibitem[{Sultan et~al.(2015)Sultan, Bethard, and Sumner}]{sultan2015dls}
Md~Arafat Sultan, Steven Bethard, and Tamara Sumner. 2015.
\newblock Dls@ cu: Sentence similarity from word alignment and semantic vector
  composition.
\newblock In {\em Proceedings of the 9th International Workshop on Semantic
  Evaluation\/}. pages 148--153.

\bibitem[{Vincent et~al.(2008)Vincent, Larochelle, Bengio, and
  Manzagol}]{vincent2008extracting}
Pascal Vincent, Hugo Larochelle, Yoshua Bengio, and Pierre-Antoine Manzagol.
  2008.
\newblock Extracting and composing robust features with denoising autoencoders.
\newblock In {\em Proceedings of the 25th international conference on Machine
  learning\/}. ACM, pages 1096--1103.

\bibitem[{Wieting et~al.(2016{\natexlab{a}})Wieting, Bansal, Gimpel, and
  Livescu}]{wieting2016charagram}
John Wieting, Mohit Bansal, Kevin Gimpel, and Karen Livescu.
  2016{\natexlab{a}}.
\newblock Charagram: Embedding words and sentences via character n-grams.
\newblock {\em arXiv preprint arXiv:1607.02789\/} .

\bibitem[{Wieting et~al.(2016{\natexlab{b}})Wieting, Bansal, Gimpel, and
  Livescu}]{wieting2016towards}
John Wieting, Mohit Bansal, Kevin Gimpel, and Karen Livescu.
  2016{\natexlab{b}}.
\newblock Towards universal paraphrastic sentence embeddings.
\newblock In {\em ICLR 2017\/}.

\bibitem[{Yessenalina and Cardie(2011)}]{yessenalina2011compositional}
Ainur Yessenalina and Claire Cardie. 2011.
\newblock Compositional matrix-space models for sentiment analysis.
\newblock In {\em Proceedings of the Conference on Empirical Methods in Natural
  Language Processing\/}. Association for Computational Linguistics, pages
  172--182.

\bibitem[{Zhou et~al.(2015)Zhou, He, Zhao, and Wu}]{zhou2015subspace}
Guangyou Zhou, Tingting He, Jun Zhao, and Wensheng Wu. 2015.
\newblock A subspace learning framework for cross-lingual sentiment
  classification with partial parallel data.
\newblock In {\em Proceedings of the international joint conference on
  artificial intelligence, Buenos Aires\/}.

\bibitem[{Zhou et~al.(2016)Zhou, Wan, and Xiao}]{zhou2016cross}
Xinjie Zhou, Xianjun Wan, and Jianguo Xiao. 2016.
\newblock Cross-lingual sentiment classification with bilingual document
  representation learning .

\end{thebibliography}

\end{document}